\documentclass[11pt]{article}
\usepackage{float}
\usepackage{inconsolata}
\usepackage[final]{acl}

\usepackage{times}
\usepackage{latexsym}

\usepackage[T1]{fontenc}

\usepackage[utf8]{inputenc}

\usepackage{microtype}

\usepackage{inconsolata}

\usepackage{graphicx}
\usepackage{tcolorbox}
\usepackage{booktabs}
\usepackage{amsfonts}
\usepackage[table]{xcolor}
\usepackage{enumitem}
\usepackage{xcolor}
\usepackage{float}
\usepackage{amsmath}
\usepackage{multirow}
\usepackage{subcaption}
\usepackage{multicol}
\definecolor{darkgreen}{rgb}{0.0, 0.5, 0.0}

\title{Router-Suggest: Dynamic Routing for Multimodal Auto-Completion in Visually-Grounded Dialogs}

\author{
\textbf{\textbf{Sandeep Mishra}$^{1}$, \textbf{Devichand Budagam}$^{1}$, Anubhab Mandal}$^{1}$, \textbf{Bishal Santra}$^{2}$, \\ \textbf{Pawan Goyal}$^{1}$, \textbf{Manish Gupta}$^{2}$\\
  $^1$IIT Kharagpur, India   $^2$Microsoft, India \\
  \texttt{\small sandeepmishraismyname@gmail.com, devichand579@gmail.com, anubhab.saie@gmail.com, } \\
  \texttt{\small bishalsantra@microsoft.com, pawangiitk@gmail.com, gmanish@microsoft.com}
}

\begin{document}
\maketitle
\begin{abstract}
Real-time multimodal auto-completion is essential for digital assistants, chatbots, design tools, and healthcare consultations, where user inputs rely on shared visual context. We introduce Multimodal Auto-Completion (MAC), a task that predicts upcoming characters in live chats using partially typed text and visual cues. Unlike traditional text-only auto-completion (TAC), MAC grounds predictions in multimodal context to better capture user intent. To enable this task, we adapt MMDialog and ImageChat to create benchmark datasets. We evaluate  leading vision-language models (VLMs) against strong textual baselines, highlighting trade-offs in accuracy and efficiency. We present \textit{Router-Suggest}, a router framework that dynamically selects between textual models and VLMs based on dialog context, along with a lightweight variant for resource-constrained environments.  Router-Suggest achieves a $2.3\times$ to $10\times$ speedup over the best-performing VLM. A user study shows that VLMs significantly excel over textual models on user satisfaction, notably saving user typing effort and improving the quality of completions in multi-turn conversations. These findings underscore the need for multimodal context in auto-completions, leading to smarter, user-aware assistants. We make our code and benchmarks publicly available\footnote{\url{https://github.com/devichand579/MAC}\label{datacodeFN}}.
\end{abstract}

\section{Introduction}
\begin{figure}[!t]
    \centering
    \includegraphics[width=\linewidth]{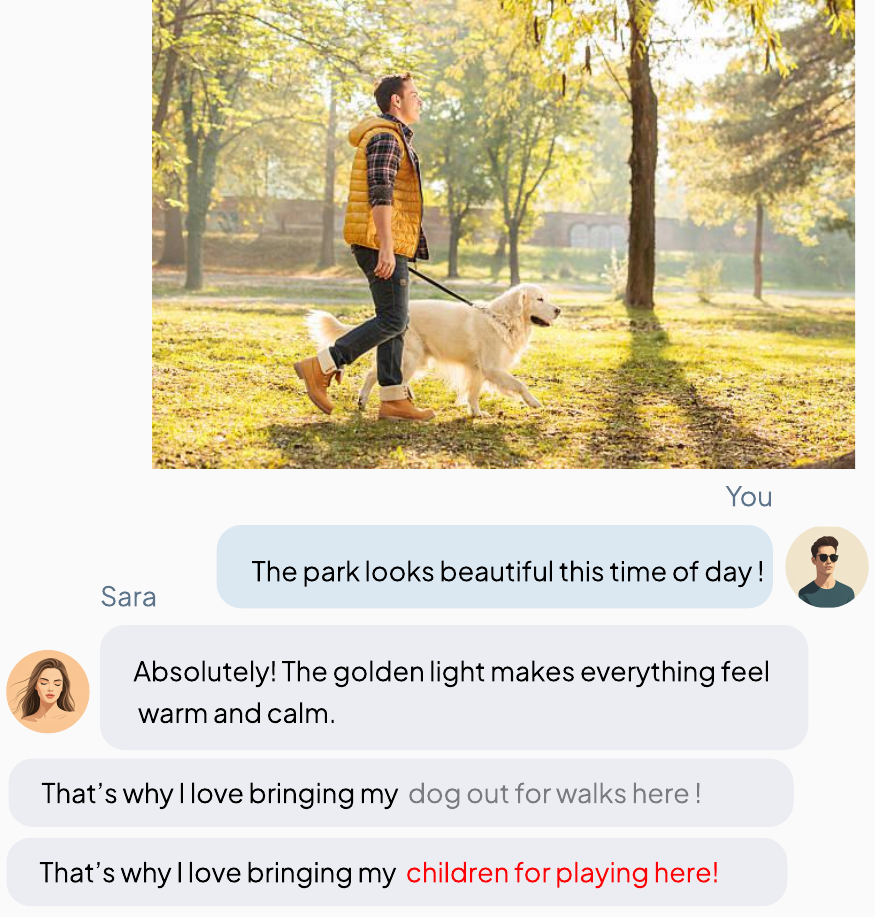}
    \caption{Example of multimodal auto-completion. Given the image context (\textit{a man walking a golden retriever in a sunlit park}) and the partial user input \textit{``That's why I love bringing my ''}, the MAC model predicts \textcolor{gray}{``dog out for walks here!''}, while a text-based TAC model incorrectly predicts \textcolor{red}{``children for playing here!''}. The MAC model prediction leverages both the textual prefix and visual context for a grounded completion.}
    \label{fig:MAC}
\end{figure}
As conversations become increasingly multimodal, the ability to predict what users will type next, while understanding both text and visuals, can transform digital assistants from reactive tools into truly intuitive partners. 
Conversational systems are increasingly used in both consumer and enterprise contexts through digital assistants, service bots, AI tools, and productivity copilots, where efficient and contextually relevant interactions are critical. Systems like ChatGPT~\cite{openai2022chatgpt} and Microsoft Copilot\footnote{\url{https://copilot.microsoft.com/}} exemplify this trend by offering intelligent, context-aware responses. Yet, as these systems evolve, user interactions increasingly include images to clarify intent, share visuals, or seek help, such as screenshots for tech support, product photos in e-commerce, design drafts in collaboration, or medical scans in telehealth. These raise new opportunities and challenges for predictive text technologies.

To streamline such interactions, inline text auto-completion (TAC) predicts user inputs in real-time using typed prefixes and dialog context. Unlike traditional query auto-completion (QAC)~\cite{bar2011context}, which presents a (dropdown) ranked list of full query suggestions, TAC offers a single completion as part of the input text field, thereby minimizing cognitive and interaction costs. However, TAC remains underdeveloped for conversational systems requiring real-time predictions in multi-turn dialogs, as most existing solutions focus on list-based QAC. For multimodal dialogues, where intent depends on both text and visuals, there exists no inline auto-completion system. Hence, we introduce Multimodal Auto-Completion (MAC), which extends TAC by using both linguistic and visual contexts to predict user input. 

MAC poses distinct challenges: (i) \textit{disambiguation under partial input}, where similar textual prefixes can warrant different completions conditioned on the image; (ii) \textit{modality alignment}, requiring the model to ground predictions in visually salient cues; and (iii) \textit{latency-efficiency trade-offs}, since vision-language inference can be substantially slower than text-only models in interactive systems.

For instance (see Figure~\ref{fig:MAC}, with an image of a man and a `golden retriever' in a park, if a user types \textit{``That's why I love bringing my ''} a TAC model might suggest \textit{``children here''} or \textit{``wife here''} ignoring the visual cue. Conversely, MAC uses the image to complete the input as \textit{``dog here''} illustrating the effectiveness of multimodal grounding. 
 
Our key contributions are as follows: 
\begin{itemize}[leftmargin=*]
\itemsep0pt
    \item \textbf{Task Definition and Benchmarking:} We define MAC as predicting inline user input from partially typed text and multimodal dialog history. To support systematic evaluation, we construct standardized benchmarks by adapting two widely used multimodal dialog datasets: MMDialog~\cite{feng-etal-2023-mmdialog} and ImageChat~\cite{shuster-etal-2020-image}, with rigorous filtering to ensure strong visual relevance.
    \item \textbf{Model Benchmarking:} We conduct a comprehensive evaluation of recent vision-language models (VLMs) like MiniCPM-V~\cite{yao2024minicpm}, PaliGemma~\cite{beyer2024paligemmaversatile3bvlm}, Qwen2-VL~\cite{yang2024qwen2technicalreport} alongside textual baselines like Most Popular Completion (MPC)~\cite{bar2011context} and Query Blazer (QB)~\cite{kang2021queryblazer} on the MAC task, highlighting key trade-offs in multimodal understanding and completion quality.
    \item \textbf{Router-Suggest:} We present a dynamic routing framework that decides, at each character, whether to use a lightweight textual model or one of the more expressive VLMs, based on the visual significance of the dialog context.
    \item \textbf{User Study:} We perform a user study to evaluate the MAC's practical effectiveness by quantifying Typing Effort Saved (TES) and user satisfaction. Results demonstrate substantial gains over text-only methods. We release our code and benchmarks\footref{datacodeFN}.
\end{itemize}
\section{Related work}
\label{sec:related}
\noindent\textbf{Query Auto-Completion (QAC):}  
QAC has long been a core component of search systems, improving efficiency and reducing query formulation effort~\cite{bast2006type}. Traditional approaches exploit signals such as popularity-based rankings~\cite{whiting2013exploring}, spatial and temporal patterns~\cite{backstrom2008spatial}, and session-level co-occurrence statistics~\cite{bar2011context}. Implementations range from classical machine learning~\cite{di2015comparing,sordoni2015hierarchical} to modern neural architectures, including LSTMs~\cite{wang2020efficient} and transformer-based models like BERT and BART~\cite{mustar2020using}.

\noindent\textbf{Text-only Auto-Completion (TAC):}  
TAC, or \textit{inline auto-completion}, also called \textit{ghosting}~\cite{ramachandran2019ghosting}, offers a single continuation within the input field, unlike QAC’s ranked suggestions. This design suits conversational contexts where dropdowns disrupt flow. Early neural methods used subword language models~\cite{kim-2019-subword} for token-level efficiency, while transformer models such as GPT-2 have been fine-tuned for next-phrase prediction in structured domains~\cite{lee-etal-2021-improving-text-auto}. More recently, reinforcement learning approaches~\cite{chitnis2024sequential,li2024ircoco} have emerged for TAC. Additional literature is provided in Appendix~\ref{sec:appendix_related_work}.

Research on dialog systems largely focuses on next-utterance prediction, whereas inline auto-completion, i.e., predicting user input mid-turn, remains underexplored. This challenge intensifies in multimodal contexts where images influence intent. Existing models prioritize full-turn responses, neglecting real-time mid-turn predictions. We introduce MAC to bridge this gap, generating grounded continuations of partial inputs using dialog history and visual context, linking full-turn response generation with real-time typing assistance in vision-language interfaces.

\section{Methods for MAC}
\label{sec:method}
\subsection{The MAC Task Definition}
\label{sec:taskDefinition}
The MAC task aims to generate a contextually appropriate continuation of a user’s partially typed input by leveraging both textual and visual dialog history. The model input comprises:  
(1) a textual prefix $p \in \mathcal{V}^{\leq T}$, representing the user’s partially typed message, where $\mathcal{V}$ is the vocabulary and $T$ is the maximum prefix length; and  
(2) a multimodal dialog history of $k$ previous utterances, $\mathcal{H}_{\text{mm}} = \{(u_1, m_1), (u_2, m_2), \dots, (u_k, m_k)\}$, where $u_i \in \mathcal{V}^{l_i}$ is a prior utterance of length $l_i \leq T$ and $m_i \in \mathcal{M}$ is an optional associated modality such as an image.

The model outputs a textual continuation $c$ such that the concatenated sequence $[p; c]$ forms a fluent and contextually coherent  message with respect to the multimodal dialog context $\mathcal{H}_{\text{mm}}$. We learn model parameters $\theta$ that maximize the conditional likelihood of $c$ given the prefix and multimodal context:
\[
    \theta^* = \arg\max_{\theta} \, P(c \mid p, \mathcal{H}_{\text{mm}}; \theta)
\]
At inference, given a new prefix $p$ and context $\mathcal{H}_{\text{mm}}$, the model generates a prediction $\hat{c}$ via:
\[
    \hat{c} = \arg\max_{c} \, P(c \mid p, \mathcal{H}_{\text{mm}}; \theta^*)
\]

This formulation enables real-time auto-completion during multimodal interactions, improving typing efficiency and coherence in visually grounded conversations.

\subsection{Benchmark Construction for MAC Evaluation}
\label{sec:benchmark}
Progress on multimodal auto-completion has been limited by the absence of standardized benchmarks. Existing multimodal dialog datasets rarely emphasize visual context as a key driver of user intent. To address this, we adapt two prominent multimodal dialog datasets: MMDialog~\cite{feng-etal-2023-mmdialog} and ImageChat~\cite{shuster-etal-2020-image} for the MAC task.

\begin{table}[t]
\centering
\scriptsize
\begin{tabular}{lrrrrr}
\hline
\rowcolor{gray!20}
\textbf{Dataset} & \textbf{Split} & \textbf{\# Dialogs} & \textbf{Avg Uttr Len} & \textbf{Avg \# Uttr} \\
\hline
\multirow{2}{*}{MMDD}       
    & Train & 13,182 & 51.81 & 11.97 \\
    \cline{2-5}
    & Test  &   893  & 50.96 & 12.80 \\
\hline
\multirow{2}{*}{ImageChat}  
    & Train & 186,724 & 49.32 & 1.91 \\
    \cline{2-5}
    & Test  &  9,994  & 49.44 & 3.00 \\
\hline
\end{tabular}

\caption{MAC Benchmark Dataset statistics after preprocessing. 
Length is measured in characters.}
\label{tab:dataset_stats}
\end{table}
We utilize GPT-4V~\cite{openai2023gpt4v} to filter datasets, selecting dialogs where images are essential for predicting the user's next input, ensuring visual grounding. We focus on single-image conversations to allow accurate visual relevance assessment without hallucinations. MMDialog (MMDD)~\cite{feng-etal-2023-mmdialog} includes domain-specific conversations enhanced with visuals like movie posters and scene stills; we select cases where images significantly influence dialog flow. ImageChat~\cite{shuster-etal-2020-image} offers open-domain conversations linked to images. 

Following the filtering and formatting steps, the curated versions of MMDialog and ImageChat form robust MAC benchmarks. Table~\ref{tab:dataset_stats} summarizes the key statistics: MMDialog features longer dialogs with more utterances per conversation, while ImageChat contains shorter, image-grounded exchanges. Additional details appear in Appendix~\ref{app:benchmark}.
\begin{figure*}[t]
    \centering
    \includegraphics[width=\linewidth]{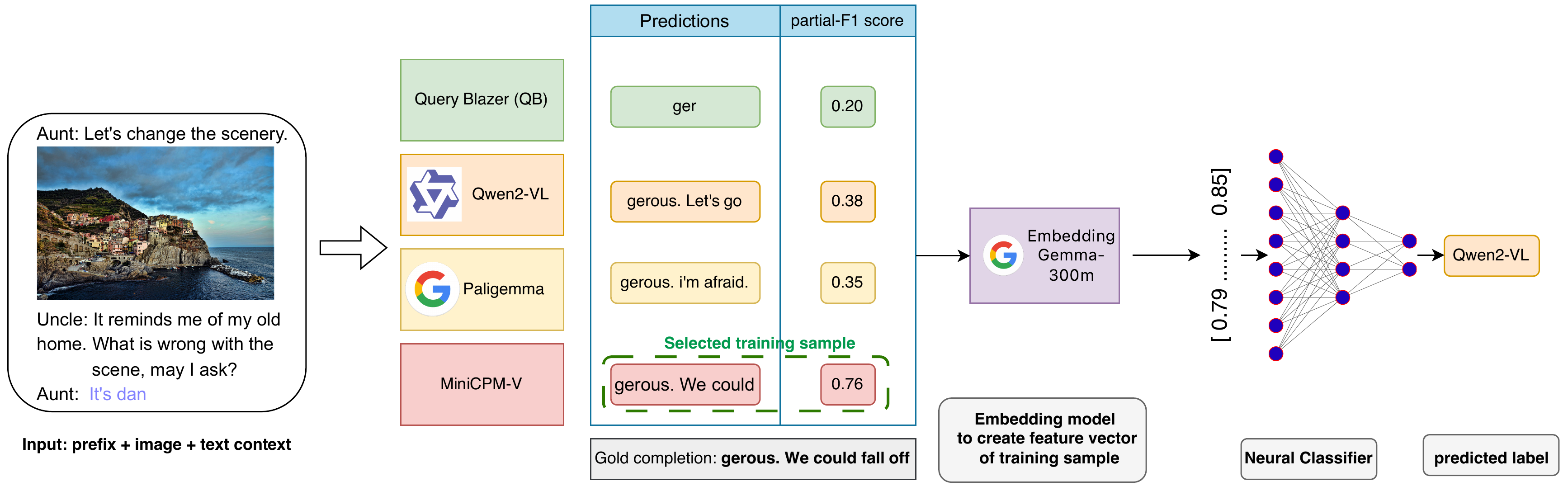}
    \caption{During router training, VLMs receive the entire input context, while the textual QB model only uses the prefix. We calculate partial-F1 scores of predictions to determine the gold label. Further, we generate a feature vector for the input prefix of the training sample using EmbeddingGemma-300m for training the neural classifier.}
    \label{fig:router}
\end{figure*}
\subsection{Models for the MAC Task}
We benchmark both textual models and VLMs, ranging from traditional retrieval-based approaches to modern VLMs, for the MAC task. Appendix~\ref{sec:appendix_baselines} lists additional information about these models.

\textbf{Textual Models:} These include trie-based methods such as \textit{Most Popular Completion (MPC)}~\cite{bar2011context}, \textit{MPC++}~\cite{bar2011context} and n-gram based model \textit{QueryBlazer (QB)}~\cite{kang2021queryblazer}.

\textbf{Vision Language Models (VLMs):} These include MiniCPM-V~\cite{yao2024minicpm}, PaliGemma (3B)~\cite{beyer2024paligemmaversatile3bvlm} and Qwen2-VL~\cite{wang2024qwen2vlenhancingvisionlanguagemodels}.
\subsection{The Proposed Router-Suggest Framework}
Textual models and VLMs vary significantly in terms of their latency and accuracy. To balance these trade-offs, we present \textit{Router-Suggest}, which adaptively selects the optimal model per prefix. we frame routing as a classification problem, where a lightweight neural router predicts the best model based on input complexity. The average system latency with a router configuration with $n$ MAC models can be computed as:
\[ L_{\mathrm{total}} = L_{\mathrm{Router}} \\
+ \sum_{i=1}^{n} p_i \cdot L_i \]
where,
$p_i$ is the probability of triggering the $i$-th MAC model. We employ a lightweight neural classifier as a router to minimize the router's latency overhead, i.e., $L_{\mathrm{Router}} \approx 0$. For router training (See Fig.~\ref{fig:router}), for each training (prefix, completion) sample, we use 768D EmbeddingGemma-300m~\cite{vera2025embeddinggemmapowerfullightweighttext} representations of input prefixes as features. To train the router, we obtain the ground truth optimal model for each sample as follows. First, we generate completions for an input prefix using all the models. The model with the highest partial-F1 score against the true completion is selected as the ground truth optimal model.

To incorporate latency-awareness, we perform cost-sensitive training of the router. 
For $C$ candidate models (and hence number of classes for router) and a batch of $N$ samples, let $p_s^m$ denote the predicted probability for model $m\in [1,c]$ and sample $s\in [1,N]$, and $c^m$ its cost  proportional to its average latency.
Let $y_s$ denote the true class label for sample $s$. Then we compute the cross entropy loss for the batch as:
\[
\mathcal{L}_{\text{CE}} = -\frac{1}{N} \sum_{i=1}^{N} \log p_s^{y_s}
\]

For each sample $s$, the expected cost is the probability-weighted average of per-class costs:
\[
\mathbb{E}_{p_s}[\text{cost}] = \sum_{m=1}^{C} p_{s}^m c^m
\]
Averaged across the batch:
\[
\mathcal{L}_{\text{Cost}} = \frac{1}{N} \sum_{s=1}^{N} \sum_{m=1}^{C} p_s^m c^m
\]
The overall loss $\mathcal{L}$ combines accuracy and cost-awareness in a single objective.
$\mathcal{L}_{\text{CE}}$ encourages correct classification, while 
$\mathcal{L}_{\text{Cost}}$ penalizes predictions with higher expected costs.
\[
\mathcal{L} = (1 - \lambda)\,\mathcal{L}_{\text{CE}}
+ \lambda\,\mathcal{L}_{\text{Cost}}
\]
The trade-off parameter $\lambda$ enables a controlled compromise between accuracy and cost efficiency. The routing framework is model-agnostic, integrating the text-based TAC and MAC models with different latency-accuracy trade-offs. This ensures efficient, real-time deployment of multimodal completion systems with high completion quality. At test time, we select the model having the highest probability predicted by the router. 

\section{Experiments and Results}
\label{sec:exp_res}
\tabcolsep2pt
\begin{table}[t]
\centering
\scriptsize
\begin{tabular}{l|l|ccccccc}
\hline
\textbf{Type} & \textbf{Model} & \textbf{TR} & \textbf{SM} & \textbf{PR-P} & \textbf{PR-R} & \textbf{PR-F1} & \textbf{|Pred|} & \textbf{TES} \\
\hline
\rowcolor{gray!20} \multicolumn{9}{c}{\textbf{MMDD}} \\
\hline
\multirow{4}{*}{Textual} 
 & MPC       & 0.1991 & 0.0000 & 0.0782 & 0.0676 & 0.0725 & \textbf{40.6} & 0.0015 \\
 & MPC++     & 0.5651 & 0.0332 & 0.1831 & 0.1303 & 0.1525 & 29.4 & 0.0430 \\
 & QB        & 0.9220 & 0.0426 & \textbf{0.3498} & 0.1287 & 0.1892 & 8.9  & 0.1724 \\
\noalign{\vskip 1pt}\arrayrulecolor{black}\specialrule{.5pt}{0pt}{0pt}
\multirow{3}{*}{VLMs} 
 & MiniCPM-V & \textbf{0.9898} & \textbf{0.1182} & 0.3362 & \textbf{0.2423} & \textbf{0.2800} & 21.1 & \textbf{0.2136} \\
 & PaliGemma & 0.9880 & 0.0972 & 0.2896 & 0.2145 & 0.2470 & 20.3 & 0.2030 \\
 & Qwen2-VL  & 0.9891 & 0.1034 & 0.2950 & 0.2223 & 0.2532 & 18.8 & 0.1844 \\
\hline

\rowcolor{gray!20} \multicolumn{9}{c}{\textbf{ImageChat}} \\
\hline
\multirow{4}{*}{Textual} 
 & MPC         & 0.2749 & 0.0007 & 0.1120 & 0.0685 & 0.0845 & \textbf{27.7}  & 0.0030 \\
 & MPC++       & 0.6728 & 0.0341 & 0.2080 & 0.1202 & 0.1523 & 17.3  & 0.0371 \\
 & QB          & 0.9604 & 0.0373 & 0.3065 & 0.1225 & 0.1755 & 5.9   & 0.0955 \\
\noalign{\vskip 1pt}\arrayrulecolor{black}\specialrule{.5pt}{0pt}{0pt}
\multirow{3}{*}{VLMs} 
 & MiniCPM-V   & \textbf{0.9892} & \textbf{0.0715} & \textbf{0.3128} & \textbf{0.2205} & \textbf{0.2586} & 16.1 & 0.1246 \\
 & PaliGemma   & 0.9881 & 0.0616 & 0.2850 & 0.1996 & 0.2348 & 16.7 & 0.1148 \\
 & Qwen2-VL    & 0.9889 & 0.0577 & 0.2931 & 0.1971 & 0.2356 & 16.2 & \textbf{0.1422} \\
\hline
\end{tabular}
\caption{Performance metrics on \textbf{unseen prefixes} of the MMDD (top) and ImageChat (bottom), organized by type (Textual vs. VLMs). |Pred|=Avg Pred Len. TES is calculated relative to ground truth completions.}
\label{tab:mmdd_imgchat_scores}
\end{table}
\subsection{Evaluation Metrics}
Standard NLG metrics like BLEU~\cite{papineni2002bleu}, ROUGE~\cite{lin2004rouge}, and METEOR~\cite{banerjee2005meteor} are unsuitable for MAC tasks, which require inline continuation of user input. These metrics focus on sequence overlap, but MAC needs accuracy in continuing user input to avoid cognitive load and ensure acceptance. Traditional QAC metrics such as top-$k$ accuracy or Mean Reciprocal Rank (MRR) assume a ranked list of suggestions, which is incompatible with the single, inline nature of MAC. These approaches also fail to account for the real-time aspect of interaction, when and how often suggestions are triggered.

To address these limitations, we utilize a set of MAC-specific metrics from~\cite{mishra2025chatghostingcomparativestudymethods}, including Trigger Rate (TR), Syntactic Match (SM), Partial Recall (PR-R), Partial Precision (PR-P), Partial F1 (PR-F1), and Typing Effort Saved (TES). These metrics provide a precise assessment of the usability, accuracy, and efficiency of real-time multimodal chat system completions.

Let $s_i$ be the model's suggestion for instance $i$, $g_i$ be the ground truth continuation for instance $i$ and $N$ denote the number of utterances in the evaluation dataset.
\begin{itemize}[leftmargin=*]
    \item  \textbf{Syntactic Match (SM):} 
    SM measures the percentage of model-generated completions that exactly match the ground truth continuation. A completion is considered a syntactic match if it is identical to the reference output when suggestions are shown.
\[ \text{SM} = \frac{1}{N} \sum_{i=1}^{N} \mathbb{I}(s_i = g_i) \]
where $\mathbb{I}(\cdot)$ is the indicator function that returns 1 if the condition is true, and 0 otherwise.
    \item \textbf{Partial Recall (PR-R):}
PR-R quantifies the average percentage of ground truth characters that overlap with the predicted completion, starting from the beginning. It reflects how much of the true continuation the model successfully recovered as a prefix.
\[ 
\text{Recall}_p = \frac{1}{N} \sum_{i=1}^{N} \frac{\text{len}(\text{prefix\_match}(s_i, g_i))}{\text{len}(g_i)} 
\]
where $\text{prefix\_match}(s_i, g_i)$ returns the longest common prefix between $s_i$ and $g_i$. 
    \item \textbf{Partial Precision (PR-P):}
PR-P quantifies the average percentage of predicted characters that overlap with the ground truth continuation, starting from the beginning. It reflects how much of the predicted completion is actually correct as a prefix.
\[ \text{Precision}_p = \frac{1}{N} \sum_{i=1}^{N} \frac{\text{len}(\text{prefix\_match}(s_i, g_i))}{\text{len}(s_i)} \]
    \item \textbf{Trigger Rate (TR)}: TR measures how frequently a suggestion is shown to the user, based on a predefined confidence threshold. It is calculated as the ratio of the number of times a suggestion was triggered to the total number of characters typed by the user.
    \[
    \text{TR} = \frac{1}{N} \sum_{i=1}^{N} \frac{\text{\# suggestions triggered}_i}{\text{\# total characters typed}_i}
    \]
    \item \textbf{Typing Effort Saved (TES)}: TES measures the proportion of ground truth characters saved, i.e., the overlap between prediction and target continuation. TES can be interpreted as a normalized keystroke saving rate across the entire dataset.
    \[
    \text{TES} =\frac{1}{N} \sum_{i=1}^{N} \left(1 - \frac{\text{\# characters actually typed}_i}{\text{total utterance length}_i}\right)
    \]
\end{itemize}
These metrics assess several aspects of the MAC task: \emph{accuracy} (assessed through PR-P, PR-R and the partial-F1, which represents the harmonic mean of PR-P and PR-R), \emph{usability} (via TES and TR), and \emph{syntactic fluency} (via SM). Collectively, they enable a more comprehensive understanding of model behavior than traditional metrics and are essential for benchmarking MAC systems.

\subsection{Finetuning Setup}
\label{sec:finetuning}
We perform two pre-processing steps (unrolling and splitting) on the dialog datasets to format them into the standard structure desired: \textit{context + image + prefix + completion}. In the unrolling step, the dialog is progressively built by appending each utterance one at a time, resulting in an increasingly rich context.
In the splitting step, the entire conversation is preserved up to the penultimate utterance. The last utterance is then randomly divided into two segments: the first serves as the prefix, and the second becomes the target completion to be predicted.

We trained our text models using default settings, closely following QB~\cite{kang2021queryblazer}, which includes a 4,096-token vocabulary that covers 99.95\% of characters. Subsequently, an 8-gram language model was constructed with pruning. Models utilizing both MPC~\cite{bar2011context} and MPC++~\cite{bar2011context} were implemented with their standard configurations. For the VLM-based models, we conducted training over 5 epochs, using a batch size of 8 per device and a learning rate of 0.0001. This process employed mixed-precision (\texttt{FP16}) training. LoRA adapters, with a rank of 8, were incorporated into all linear layers and subjected to a 0.05 dropout rate. Throughout this, we maintained the base model in a frozen state, updating only the LoRA parameters.

\subsection{Performance on MAC Benchmarks}
Table~\ref{tab:mmdd_imgchat_scores} reveals a clear performance gap between text models and VLMs on unseen prefixes across both MMDD and ImageChat datasets. Text models collapse in MMDD, with MPC showing nearly zero Syntactic Match $(SM = 0)$ and $TES~(0.0015)$, indicating severe overfitting. Even the enhanced MPC++ offers limited gains, while QB generalizes modestly but still deteriorates in multimodal contexts. In contrast, VLMs maintain consistently high Trigger Rates $(TR \approx 0.99)$ and stable PR metrics, leveraging multimodal grounding for robust contextual completions. MiniCPM-V achieves the best overall $TES~(0.2136)$ and balanced PR scores while generating shorter, more efficient completions ($\approx 18$-$22$ characters) compared to verbose outputs from text models (e.g., MPC $|Pred| = 40.6$). 

On ImageChat, the gap narrows as text models degrade less sharply, but VLMs still outperform, sustaining higher TES and smoother precision–recall trade-offs. Overall, VLMs demonstrate superior generalization and adaptability in unseen multimodal scenarios. Please see Appendix~\ref{sec:appendix_seen} for results on seen prefixes on both benchmarks.

\subsection{Evaluation of Router-Suggest}
Table~\ref{tab:router_general} presents the latency-performance tradeoff of individual models alongside Router-Suggest. The absolute latencies for all VLMs are determined through inference using vLLM~\cite{kwon2023efficient} as the inference engine, applied to a representative dataset consisting of prefixes from both MMDD and ImageChat. We conducted a joint hyperparameter and architectural search for router configurations across various $\lambda$ (See Fig.~\ref{fig:router_tradeoff}) to optimize performance and latencies, as detailed in Appendix~\ref{sec:appendix_router}. 

\begin{table}[!b]
\centering
\scriptsize
\setlength{\tabcolsep}{2pt}
\begin{tabular}{lcccccc}
\hline
\multirow{2}{*}{\textbf{Model}} & \multicolumn{3}{c}{\textbf{MMDD}} & \multicolumn{3}{c}{\textbf{ImageChat}} \\
\cline{2-4}\cline{5-7}
& \textbf{PR-F1} & \textbf{SM} & \textbf{Time (s)$\downarrow$} & \textbf{PR-F1} & \textbf{SM} & \textbf{Time (s)$\downarrow$} \\
\hline
\rowcolor{gray!20}
\multicolumn{7}{c}{\textbf{Individual Models}}\\
\hline
MiniCPM-V       & 0.247 & 0.116 & 2.080  & \textbf{0.223} & \textbf{0.067} & 2.080  \\
PaliGemma     & 0.216 & 0.097 & 1.490  & 0.199 & 0.057 & 1.490  \\
QB            & 0.209 & 0.102 & \textbf{0.001} & 0.135 & 0.036 & \textbf{0.001} \\
Qwen2-VL      & 0.222 & 0.101 & 0.733 & 0.197 & 0.053 & 0.733 \\
\hline
\rowcolor{gray!20}
\multicolumn{7}{c}{\textbf{Router-Suggest}}\\
\hline
Router-4-L  & 0.240 & 0.110 & 0.351 & 0.212 & 0.056 & 0.966 \\
Router-4-P    & \textbf{0.281} & \textbf{0.135} & 0.832 & 0.212 & 0.056 & 0.966 \\
Router-2-L          & 0.240 & 0.109 & 0.170 & 0.196 & 0.053 & 0.288 \\
Router-2-P          &  0.261 & 0.122 & 0.271 & 0.196 & 0.053 & 0.288\\
{\color{gray}
Router-4-Max (Oracle)}& {\color{gray}0.356} & {\color{gray}0.195} & {\color{gray}--}    & {\color{gray}0.281} & {\color{gray}0.090} & {\color{gray}--} \\
\hline
\end{tabular}
\caption{Performance and latency comparison of individual models and Router-Suggest configurations across MMDD and ImageChat.}
\label{tab:router_general}
\end{table}

Router-Suggest with 4 models (QB, Qwen2-VL, PaliGemma and MiniCPM-V) needs $\sim$25GB memory on an Nvidia L40 GPU for inference. For constrained environments, we also experiment with a router configuration with just 2 models (QB, Qwen2-VL), requiring only $4GB$ GPU memory. We refer to router configurations as Router-4 and Router-2, respectively. Further, after joint hyperparameter and architecture search, we choose 2 configurations: L and P. Router-L corresponds to the hyperparameter configuration that leads to minimum latency with performance (PR-F1)  close to the best model. Router-P corresponds to the hyperparameter configuration that leads to maximum performance (PR-F1). We also compute the oracle performance of the Router-4 configuration, where the best perfroming model is always chosen for every prefix.

\begin{figure*}
    \centering
    \begin{subfigure}{0.43\textwidth}
        \centering
        \includegraphics[width=\linewidth]{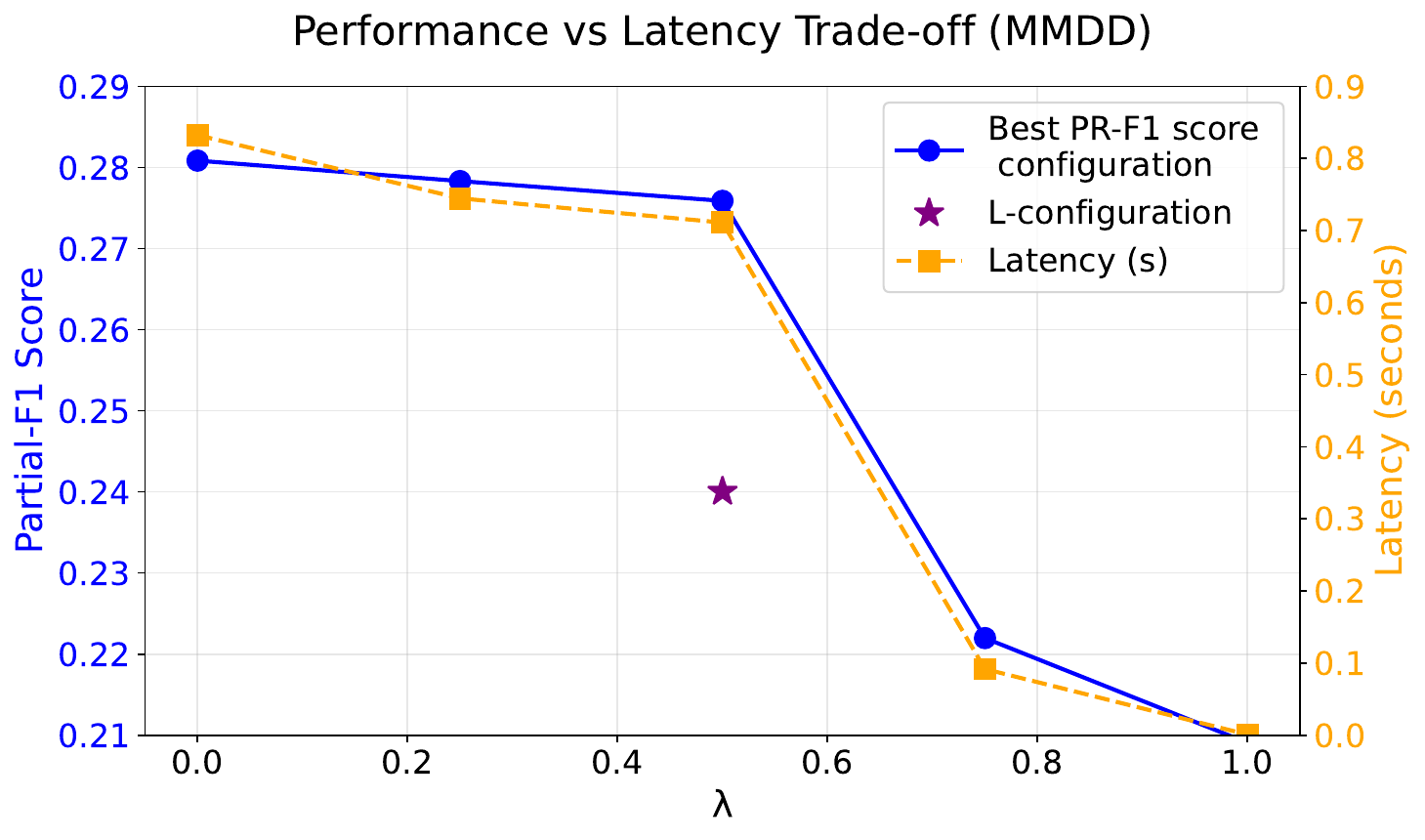}
        \caption{}
        \label{fig:subfigure1}
    \end{subfigure}%
    \begin{subfigure}{0.43\textwidth}
        \centering
        \includegraphics[width=\linewidth]{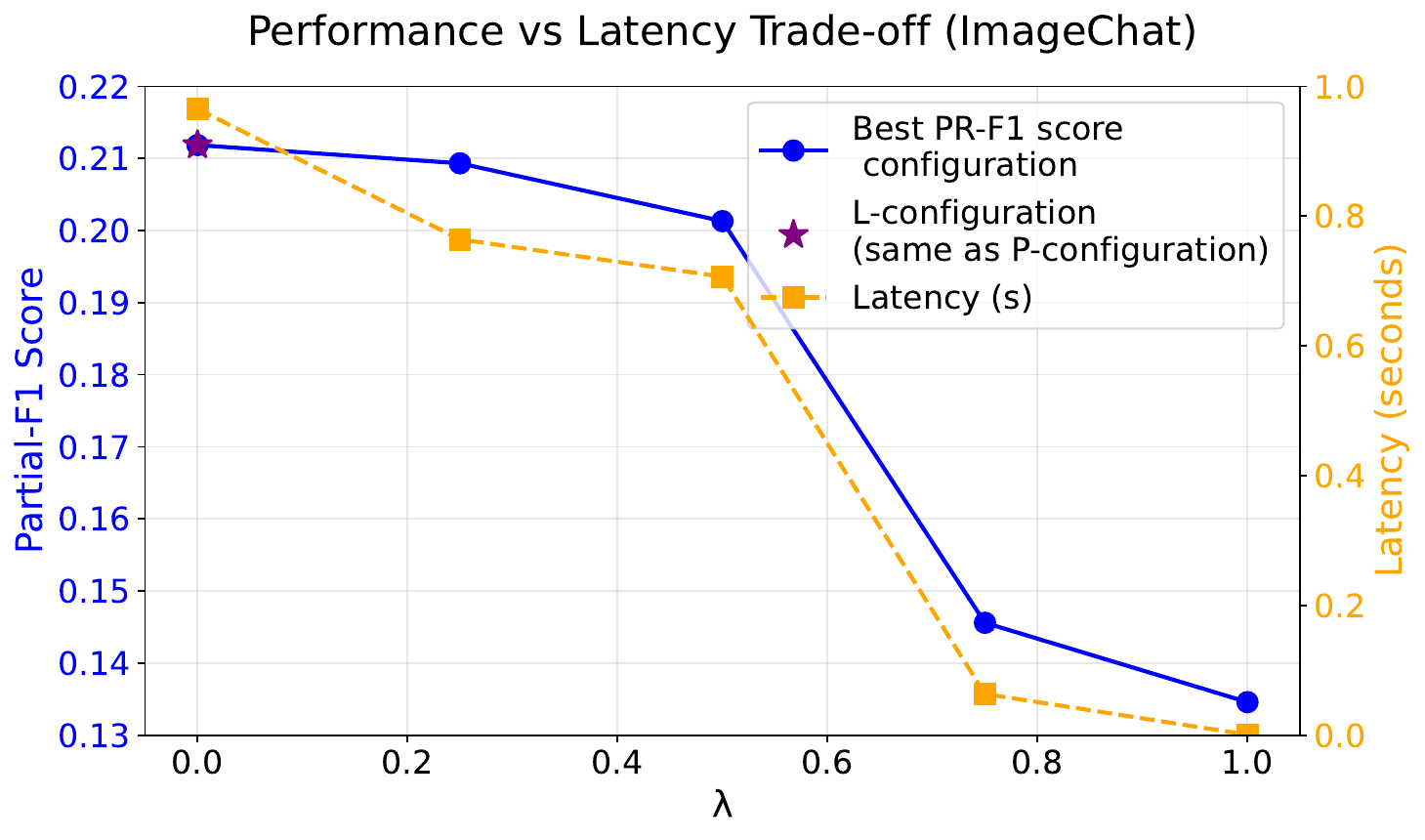}
        \caption{}
        \label{fig:subfigure2}
    \end{subfigure}
    \caption{Different router configurations for Router-4 at different $\lambda$ and their latency vs PR-F1 score tradeoff for (a) MMDD and (b) ImageChat.}
    \label{fig:router_tradeoff}
\end{figure*}
Router-4-L achieves near-competitive performance of the best-performing individual model with minimal latency, while Router-4-P offers the highest PR-F1 score. Thus, Router-Suggest models improve PR-F1 and syntactic match, reducing latency compared to high-capacity models, showcasing lightweight routing's efficiency. On MMDD, Router-4-L matched MiniCPM-V's PR-F1 score at 5$\times$ faster response time. Router-4-P achieved a PR-F1 of $0.281$, close to the $0.356$ upper bound at one-third the latency of MiniCPM-V. On ImageChat, routing maintains accuracy with minimal time overhead, highlighting scalability and practical benefits. 

Router-2-L achieves near-optimal PR-F1 compared to Qwen2-VL ($0.248$ on MMDD, $0.192$ on ImageChat) with substantially reduced latency compared to Qwen2-VL and a speedup $10\times$ compared to the best-performing model (MiniCPM-V), demonstrating effective lightweight routing. 

\section{User Study}
\begin{figure}[t]
    \centering
    \includegraphics[width=0.8\linewidth]{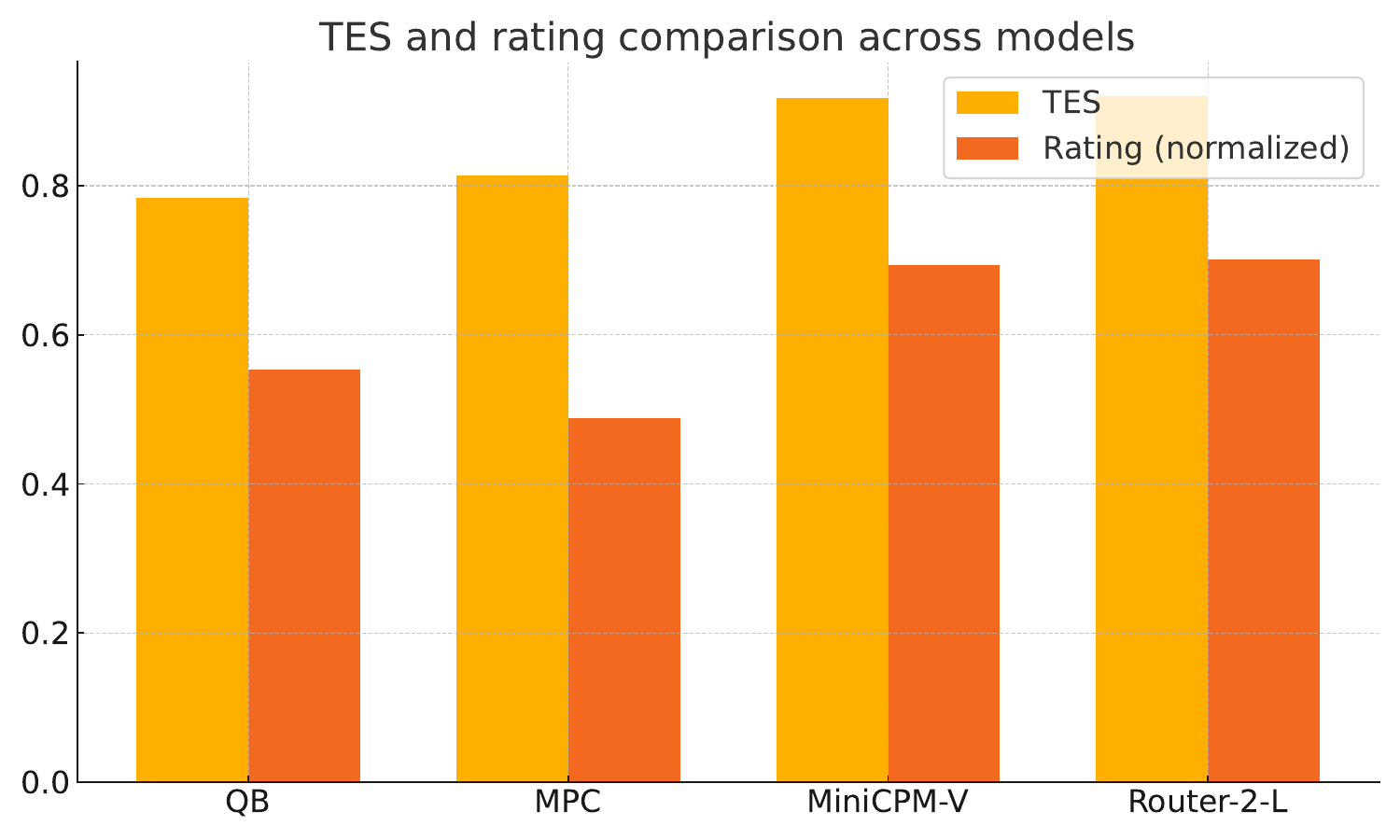}
    \caption{Comparison of mean TES and user ratings (normalized) for various models. TES is calculated relative to the final text approved by the user at the moment the rating is submitted.}
    \label{fig:user_study}
\end{figure}
We developed a platform where anonymous users can participate in completing conversations initialized from randomly selected samples of the MMDD and ImageChat datasets. During interactions, users engage with a randomly selected model (QB, MPC, or MiniCPM-V) without knowing the specific model, thus minimizing bias. Users assess the system's completion on a scale from $0$ to $9$, where $9$ represents the most satisfactory and well-aligned completion and $0$ indicates a completely unaligned, poor, or absent completion. 
TES calculation is based on the final user query at the moment the rating is submitted. Our study encompasses 190 sessions, distributed as follows: 53 with MPC, 47 with QB, 45 with MiniCPM-V and 45 with Router-2-L.

Figure~\ref{fig:user_study} illustrates a strong positive relationship between TES and user ratings across models. The visual trend confirms that as TES increases, user ratings also rise. These TES scores are significantly higher than the offline TES scores (Table~\ref{tab:mmdd_imgchat_scores}). This is expected because, in interactive settings, users often adapt their typed continuations based on the system's suggestions. As a result, the `ground truth' becomes partially influenced by the model itself, naturally inflating agreement metrics such as TES. MiniCPM-V consistently outperforms the text models, achieving the highest TES and an unnormalized user rating and router-2-L also achieved similar scores.
This demonstrates that VLMs not only achieve higher TES but also deliver a more stable and satisfying user experience than the textual counterparts.
\section{Conclusion}
\label{sec:conclusion}
We propose Multimodal Auto Completion (MAC), a novel task for predicting user input in visually grounded conversations, along with standardized benchmarks from MMDialog and ImageChat and an evaluation protocol designed for inline auto-completion. Experiments reveal textual models excel with known prefixes but struggle with new ones, whereas VLMs maintain high trigger rates and better TES and robustness in new conditions. Router-Suggest selectively engages VLMs, providing competitive partial-F1  as the best models with 2.3-10$\times$ speedup. We also provide a low-resource setup for Router-Suggest. A user study confirms TES as a reliable user satisfaction measure, aligning with subjective ratings and shows that VLM completions better meet user expectations compared to outputs from textual models. Overall, these results highlight the potential of visually grounded completions to significantly reduce typing effort and enhance perceived usefulness in practical interactive environments.

\section{Limitations}
The MAC benchmarks, adapted from MMDialog and ImageChat using GPT-4V filtering, may introduce selection bias toward visually explicit cases and lack linguistic diversity. Current datasets only cover single-image contexts, limiting generalization to real-world multimodal settings with evolving or multiple visuals. Router-Suggest, though effective in reducing latency, relies on embedding-based heuristics that may degrade under domain shift and lacks interpretability in its routing choices. 
\section{Ethical Considerations}
\label{sec:ethics}
The MAC benchmark is built using automated relevance filtering (GPT-4V) and curated public corpora, which may introduce noisy labels, annotation biases, privacy concerns, and hallucination risks. The user study relies primarily on TES and a small user pool, which may overlook key factors: TES can fail to capture subtle misinformation, cultural or demographic mismatches, and sampling choices can introduce biases that limit generalizability. Additionally, the router’s invocation patterns raise fairness and cost-allocation concerns, as it may disproportionately route certain input types or user groups to more compute-intensive MAC models, leading to unequal latency, computational cost, or quality of experience.
\bibliography{custom}

\appendix

\section{Additional Related Work}
\label{sec:appendix_related_work}
Recent work in multimodal dialog systems has focused on generating context-aware responses by integrating both visual and textual dialog history inputs.~\citet{sun-etal-2022-multimodal} proposed \textsc{Divter}, a dual-channel model that enables text or image response generation under low-resource conditions by decoupling textual and visual training.~\citet{kong-etal-2024-tiger} introduced \textsc{TIGER}, a unified transformer-based framework capable of producing text, image, or mixed-modal responses by dynamically selecting the output modality.~\citet{yoon2024bi} presented \textsc{BI-MDRG}, which incorporates visual history across dialog turns to maintain object consistency and support grounded response generation. Earlier approaches, such as \textsc{MAGIC} and \textsc{MATE}, applied transformer-based cross-modal attention mechanisms~\cite{electronics11203409} to generate visually coherent textual responses, highlighting the role of structural alignment between modalities.

\section{Benchmark Construction}
\label{app:benchmark}
\begin{figure}
    \centering
\begin{tcolorbox}[colback=gray!5, colframe=black, title=\textbf{\scriptsize Prompt Template}, fonttitle=\scriptsize, boxrule=0.5pt, left=2pt, right=2pt, top=2pt, bottom=2pt]
\scriptsize
You are a discriminator model who will decide if the following hold:
\begin{enumerate}[label*=\arabic*.]
    \item The dialog is relevant to the image.
    \item The image fits the context and is accounted for in the following utterances.
    \item The image and the dialog are coherent.
    \item The image can be used for autocompletion of following utterances.
    \item The image should not be the last utterance because it is of no use then.
\end{enumerate}
\textbf{The user will provide the dialog starting from when the image was shared and including up to 3 subsequent utterances. Carefully assess how much the image contributes to the conversation. Think through the following questions step by step before assigning a score:}

\textbf{Step-by-step Analysis:}
\begin{enumerate}[label*=\arabic*.]
    \item Provide a caption for the image (regardless of the conversation).
    \item Is the image misleading? Does it contradict or confuse the dialog?\\
    \textit{If yes, rate it lower.}
    \item Is the image completely ignored? Do the following messages continue without acknowledging it at all?\\
    \textit{If yes, rate it low.}
    \item Does the image add some relevance? Do the next messages mention something loosely connected to it, even if the dialog still makes sense without it?\\
    \textit{If yes, give a mid-range score.}
    \item Is the image clearly useful? Do the messages directly reference the image, making the conversation easier to understand?\\
    \textit{If yes, score it higher.}
    \item Is the image essential? Would the dialog be incomplete, confusing, or meaningless without it?\\
    \textit{If yes, give the highest score.}
\end{enumerate}

\textbf{Your Task:} Provide your response in valid JSON format:

\begin{tcolorbox}[colback=white!95!gray,colframe=gray!80!black]
\begin{verbatim}
<results>
{
  "caption": "<caption>",
  "answer": <score between 1-5>,
  "explanation": "<Step-by-step reasoning
                  for the score>"
}
</results>
\end{verbatim}
\end{tcolorbox}

\textbf{Scoring Scale:}
\begin{itemize}
    \item \textbf{1} → The image contradicts or misleads the dialog.
    \item \textbf{2} → The image is ignored and not acknowledged at all.
    \item \textbf{3} → The image is loosely relevant, but the dialog makes sense without it.
    \item \textbf{4} → The image adds context and is referenced, but isn’t crucial.
    \item \textbf{5} → The image is critical, and the dialog wouldn’t make sense without it.
\end{itemize}

\textbf{Important:} Justify your score with logical reasoning before assigning it.

\end{tcolorbox}
\caption{Prompt template for relevance filtering using GPT-4V.}
    \label{fig:prompt}
\end{figure}
\subsection{Relevance filtering using GPT-4V} 
\label{sec:appendix_filtering}
To ensure that images meaningfully contribute to the dialog, we employ GPT-4V~\cite{openai2023gpt4v} as an automatic discriminator to assess the relevance of each image-dialog pair, using the prompt template illustrated in Figure~\ref{fig:prompt}. Each sample is rated on a standardized 5-point scale:
\textbf{1} = Contradictory,
\textbf{2} = Ignored,
\textbf{3} = Marginally relevant,
\textbf{4} = Clearly useful,
\textbf{5} = Critical for understanding.

\begin{figure}[!t]
    \centering
    \includegraphics[width=\linewidth]{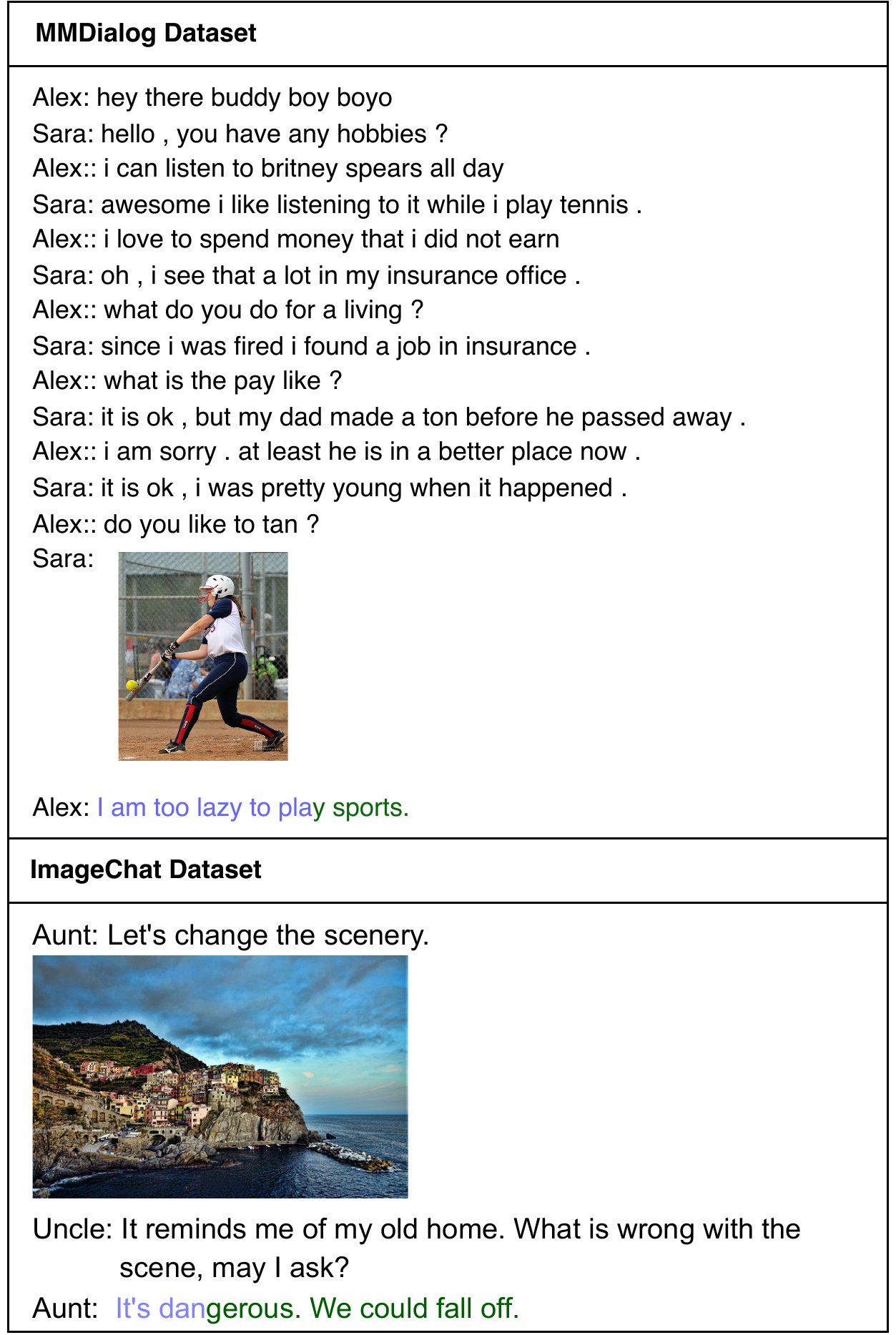}
    \caption{Two illustrative examples of MAC from the MMDialog and ImageChat datasets, where the image context significantly influences the prediction. \textcolor{blue}{Blue} indicates the input prefix provided to the MAC model, while \textcolor{darkgreen}{Green} highlights the text characters that the model is expected to predict.}
    \label{fig:dataset_ex}
\end{figure}

Only samples receiving a relevance score of \textbf{4} or \textbf{5} are retained in the final benchmark to ensure strong visual grounding and eliminate noisy or irrelevant pairs. Figure~\ref{fig:dataset_ex} illustrates examples identified as highly image-relevant by GPT-4V, highlighting the kinds of interactions that demand grounded multimodal understanding, central to the challenge of MAC. Following the filtration process, over $66\%$ of the samples were removed from the datasets.
\subsection{Formatting interleaved inputs} 
\label{sec:appendix_formating}
For models that do not natively support interleaved image-text inputs, we restructure the input to explicitly encode the position of visual content. Image embeddings are prepended to the input sequence, and a special token such as \texttt{<IMAGE>} is inserted at the corresponding turn in the dialog where the image appeared. This approach enables the model to attend to both the image features and their temporal alignment within the dialog. For example, a turn originally written as: ``\textit{User: That looks amazing!}'' would be transformed into ``\textit{User: \texttt{<IMAGE>} That looks amazing!}''

\section{Additional Details for Experiments}
\label{sec:appendix_additional_exp}
\subsection{Baseline Models}
\label{sec:appendix_baselines}
\textbf{Textual Models:} These models operate solely on textual input, without access to any visual modality. Trie-based methods such as \textit{Most Popular Completion (MPC)}~\cite{bar2011context} construct a character-level trie from historical user utterances to suggest completions based on frequency, while its extension \textit{MPC++}~\cite{bar2011context} uses a suffix trie to offer better coverage for previously unseen prefixes. N-gram-based methods like \textit{QueryBlazer (QB)}~\cite{kang2021queryblazer} rely on subword tokenization and n-gram language modeling to retrieve completions from historical logs and synthesize novel predictions.

\textbf{Vision Language Models:}
Recent advances in VLMs enable the processing of both textual and visual modalities. The models we explored include MiniCPM-V~\cite{yao2024minicpm}, a powerful 8B parameter VLM that integrates a SigLIP~\cite{zhai2023sigmoid} vision encoder with a Qwen2.5-7B language decoder. PaliGemma (3B)~\cite{beyer2024paligemmaversatile3bvlm} also employs a SigLIP vision encoder, coupled with the Gemma 2~\cite{team2024gemma} language model for text generation. Lastly, Qwen2-VL~\cite{wang2024qwen2vlenhancingvisionlanguagemodels} is a vision-language instruction-tuned variant from the Qwen2 series~\cite{yang2024qwen2technicalreport}, combining a Vision Transformer (ViT)~\cite{dosovitskiy2020image} encoder with the Qwen2 decoder to enable fine-grained, instruction-following capabilities across vision and text modalities.
\begin{table*}[!t]
\centering
\scriptsize
\begin{tabular}{l|l|ccccccc}
\hline
\textbf{Method} & \textbf{Model} & \textbf{TR} & \textbf{Syntactic Match} & \textbf{PR-Precision} & \textbf{PR-Recall} & \textbf{Partial-F1} & \textbf{Avg Pred Len} & \textbf{TES} \\
\hline
\rowcolor{gray!20} \multicolumn{9}{c}{\textbf{MMDD}} \\
\hline
\multirow{4}{*}{Text}
 & MPC       & 0.9679 & \textbf{0.7902} & \textbf{0.8066} & \textbf{0.8060} & \textbf{0.8063} & \textbf{27.5} & \textbf{0.7153} \\
 & MPC++     & 0.9679 & \textbf{0.7902} & \textbf{0.8066} & \textbf{0.8060} & \textbf{0.8063} & \textbf{27.5} & \textbf{0.7153} \\
 & QB        & 0.9474 & 0.2355 & 0.5508 & 0.3213 & 0.4064 & 12.1 & 0.3725 \\
\noalign{\vskip 1pt}\arrayrulecolor{black}\specialrule{.5pt}{0pt}{0pt}
\multirow{3}{*}{VLMs} 
 & MiniCPM-V & 0.9898 & 0.1349 & 0.3505 & 0.2632 & 0.3007 & 22.3 & 0.2352 \\
 & PaliGemma & 0.9880 & 0.1179 & 0.3138 & 0.2381 & 0.2707 & 20.0 & 0.2357 \\
 & Qwen2-VL  & \textbf{0.9902} & 0.1112 & 0.3016 & 0.2279 & 0.2596 & 19.9 & 0.2097 \\
\hline

\rowcolor{gray!20} \multicolumn{9}{c}{\textbf{ImageChat}} \\
\hline
\multirow{4}{*}{Text}
 & MPC         & 0.9497 & \textbf{0.2892} & 0.4559 & 0.4723 & 0.4639 & 13.7  & \textbf{0.2688} \\
 & MPC++       & 0.9497 & \textbf{0.2892} & 0.4559 & 0.4723 & 0.4639 & 13.7  & \textbf{0.2688} \\
 & QB          & 0.9741 & 0.2094 & \textbf{0.5053} & 0.4404 & 0.4708 & 8.2   & 0.2444 \\ 
\noalign{\vskip 1pt}\arrayrulecolor{black}\specialrule{.5pt}{0pt}{0pt}
\multirow{3}{*}{VLMs}
 & MiniCPM-V   & \textbf{0.9958} & 0.2100 & 0.4611 & \textbf{0.5010} & 0.4802 & 14.4 & 0.2552 \\
 & PaliGemma   & 0.9875 & 0.2020 & 0.4694 & 0.4924 & \textbf{0.4806} & \textbf{14.7} & 0.3021 \\
 & Qwen2-VL    & 0.9945 & 0.1699 & 0.4323 & 0.4617 & 0.4465 & \textbf{14.7} & 0.2464 \\
\hline
\end{tabular}
\caption{Performance metrics on \textbf{seen prefixes} of the MMDD (top) and ImageChat (bottom) test sets, organized by model type (Text vs. VLMs).}
\label{tab:mmdd_imgchat_seen}
\end{table*}
\subsection{Performance of MAC Benchmarks on Seen prefixes}
\label{sec:appendix_seen}
On seen prefixes (See Table~\ref{tab:mmdd_imgchat_seen}), textual models achieve their strongest performance, with MPC and MPC++ reaching very high syntactic and semantic alignment on MMDD 
($\text{SM} = 0.79$, $\text{F1} = 0.81$, $\text{TES} = 0.72$), indicating strong memorization and a close fit to training distributions. 
VLMs, while showing lower syntactic precision ($\text{F1} \approx 0.27$--$0.30$), maintain consistent trigger rates 
($\text{TR} \approx 0.99$) and balanced completion lengths, reflecting stable yet less overfitted behavior. 
In ImageChat, both model families perform comparably, with VLMs (MiniCPM-V, PaliGemma) matching or slightly surpassing textual models in Partial-F1 
($\approx 0.48$). Overall, textual models dominate on seen data through memorization, whereas VLMs achieve similar precision with greater contextual grounding.

\subsection{Additional Details of Router-Suggest}
\label{sec:appendix_router}
We performed joint hyperparameter and architecture search using random sampling over a structured search space, combining both network topology and training parameters. 
Each configuration was trained using a fixed batch size of 256 and dropout rate of 0.2. 
For every trade-off parameter $\lambda \in \{0.0, 0.25, 0.5, 0.75, 1.0\}$, 
we executed 50 random trials, totaling 250 experiments for each dataset.

\begin{table}[h]
\centering
\scriptsize
\setlength{\tabcolsep}{5pt}
\begin{tabular}{ll}
\hline
\textbf{Parameter} & \textbf{Search Space} \\
\hline
Hidden dimensions & 
\begin{tabular}[c]{@{}l@{}}
$[128]$, $[256]$, \\
$[128, 64]$, $[256, 128]$, $[512, 256]$, $[64, 32]$, \\
$[256, 128, 64]$, $[512, 256, 128]$
\end{tabular} \\
Epochs & \{50, 100\} \\
Learning rate & \{1e$^{-4}$, 5e$^{-4}$, 1e$^{-3}$\} \\
$\lambda$ & \{0.0, 0.25, 0.5, 0.75, 1.0\} \\
Batch size & 256 (fixed) \\
Dropout & 0.2 (fixed) \\
\hline
\end{tabular}
\caption{Search space for architecture and hyperparameter tuning. Each $\lambda$ setting was tuned independently using random search.}
\label{tab:search_space}
\end{table}

The scoring function balanced accuracy and latency using a weighted objective:
\[
\text{Score} = (1 - \lambda) \times \text{Accuracy} + \lambda \times \text{Cost},
\]
where cost values were normalized by the maximum observed latency ($\text{max cost} = 2.0891$ for \textit{MiniCPM-V}). 
This formulation ensured fair comparison across trade-off settings, allowing selection of the highest-scoring model overall.
\end{document}